\documentclass[final,5p,times,authoryear]{elsarticle}

\usepackage{amssymb}
\usepackage{algorithm}
\usepackage{algpseudocode}
\usepackage{graphicx}
\usepackage[inkscapelatex=false]{svg}
\usepackage{pdfpages}

\newcommand{\ie}{{\emph{i.e.}}, }

\usepackage{subfigure}
\usepackage{amsmath}
\usepackage{xcolor}
\usepackage{tabularx} 
\usepackage{multirow}
\usepackage{multicol}
\usepackage{booktabs} 
\usepackage{pbox}
\usepackage{amsfonts}
\usepackage{graphicx}
{\begin{list}               %    with flush left bullets.
    {$\bullet$ \hfill}{
        \setlength{\leftmargin}{\parindent}
        \setlength{\parsep}{0.04\baselineskip}
        \setlength{\itemsep}{0.5\parsep}
        \setlength{\labelwidth}{\leftmargin}
        \setlength{\labelsep}{0em}}
    }
{\end{list}}

\providecommand{\cref}[1]{Chapter~\ref{#1}}
\providecommand{\sref}[1]{Section~\ref{#1}}

\providecommand{\norm}[1]{\lVert#1\rVert}

\renewcommand{\vec}[1]{\ensuremath{\boldsymbol{#1}}}
\providecommand{\mat}[1]{\ensuremath{\boldsymbol{#1}}}

\providecommand{\calC}{\mathcal{C}}
\providecommand{\calD}{\mathcal{D}}

\providecommand{\calL}{\mathcal{L}}

\providecommand{\calY}{\mathcal{Y}}

\providecommand{\mF}{\mat{F}}

\providecommand{\mI}{\mat{I}}

\providecommand{\mM}{\mat{M}}

\providecommand{\vf}{\vec{f}}

\providecommand{\vp}{\vec{p}}

\providecommand{\vt}{\vec{t}}

\providecommand{\vv}{\vec{v}}

\providecommand{\vy}{\vec{y}}

\usepackage{url}

\journal{Neurocomputing}

\begin{document}

\begin{frontmatter}

\title{Learning from Multimodal Pseudo-Labels for Robust Open-Vocabulary \\ Instance and Panoptic Segmentation}

\author[1]{Duy Tran Thanh}
\ead{duy.tranthanh@seoultech.ac.kr}

\author[2]{Yeejin Lee}
\ead{yeejinlee@seoultech.ac.kr}

\author[3]{Byeongkeun Kang\corref{cor1}}
\ead{byeongkeunkang@cau.ac.kr}
\cortext[cor1]{Corresponding author.}

\affiliation[1]{organization={Department of Electronic Engineering, Seoul National University of Science and Technology},
            addressline={232 Gongneung-ro, Nowon-gu}, 
            city={Seoul},
            postcode={01811}, 
            country={South Korea}}
\affiliation[2]{organization={Department of Electrical and Information Engineering, Seoul National University of Science and Technology},
            addressline={232 Gongneung-ro, Nowon-gu}, 
            city={Seoul},
            postcode={01811}, 
            country={South Korea}}
\affiliation[3]{organization={School of Electrical and Electronics Engineering, Chung-Ang University},
            addressline={84 Heukseok-ro, Dongjak-gu}, 
            city={Seoul},
            postcode={06974}, 
            country={South Korea}}

\begin{abstract}
This work addresses the challenge of open-vocabulary instance segmentation (OVIS) and open-set panoptic segmentation (OSPS), which aim to recognize both predefined and unseen object categories without exhaustive human annotations. Existing methods often suffer from noisy pseudo-masks, limited visual-textual grounding, and difficulty handling synonyms or out-of-vocabulary (OOV) words. To overcome these challenges, we propose a multimodal framework that leverages pre-trained vision-language models for automatic pseudo-label generation, CLIP-guided synonym filtering, and GPT-based caption reconstruction. In our target-vocabulary-assisted pseudo-labeling setting, the framework first constructs pseudo segmentation masks, descriptive captions, and semantically aligned synonym sets using Grounded SAM, LLaVA, and CLIP, providing multimodal supervision without manual annotation. We then enhance visual-textual alignment through three complementary training objectives: an extended grounding loss that incorporates visually grounded synonyms, a semantic consistency loss, and a generative caption reconstruction loss. Extensive experiments on the COCO dataset demonstrate that the proposed method consistently outperforms previous state-of-the-art approaches under this protocol, achieving substantial improvements on both OVIS and OSPS benchmarks.
\end{abstract}


\end{frontmatter}

\section{Introduction}
\label{sec:intro}
Instance segmentation is a fundamental task in computer vision that aims to simultaneously detect, segment, and classify individual object instances, supporting a wide range of applications in robotics, autonomous driving, surveillance, and medical imaging. Although deep neural networks have achieved significant advancements in this domain, most existing approaches remain limited to detecting and segmenting objects from a predefined set of categories and require expensive pixel-level annotations for training~\citep{Tran2024MSTA3D}. To mitigate these challenges, recent studies have explored zero-shot and weakly supervised segmentation frameworks that aim to recognize novel categories with minimal human supervision~\citep{Kim2025Generalized, PRLwsss2025, Hoang_Lee_Kang_2025, biertimpel2021prior}. However, weakly supervised methods still require image-level or bounding-box annotations for all categories~\citep{PRLwsss2025}, whereas zero-shot approaches~\citep{Kim2025Generalized, bansal2018zero, zheng2021zero} rely only on textual embeddings such as word vectors. Nevertheless, zero-shot methods are typically constrained by limited generalization ability and often struggle to capture fine-grained visual details when applied to complex real-world datasets.

To overcome these limitations, researchers have explored open-vocabulary instance segmentation (OVIS) and open-set panoptic segmentation (OSPS), which integrate large-scale vision-language models (VLMs) with flexible semantic representations \citep{zareian2021open, huynh2022open, vs2023mask, wu2023betrayed, neucom2026_1, neucom2026_2, neucom2026_3, neucom2026_4, neucom2025_1, neucom2025_2, neucom2025_3, Liu2025Physically}. In these frameworks, models are trained with strong supervision for base categories and weak supervision from image-caption pairs for novel ones, enabling broader generalization to unseen classes.

XPM~\citep{huynh2022open} is one of the earliest frameworks, employing a teacher-student paradigm in which the teacher model generates pseudo-masks by aligning the visual features of object regions with word embeddings extracted from image captions. The student model is then trained on these pseudo-labels while estimating annotation noise to improve robustness against imperfect supervision. However, such methods typically rely on strong supervision for base classes and weak supervision from image-caption pairs for novel ones, leading to a bias toward base categories. To alleviate this issue, Mask-free OVIS~\citep{vs2023mask} removes the need for manual annotations by training a Mask R-CNN model with pseudo-masks generated by pre-trained vision-language models. Despite these advances, both XPM~\citep{huynh2022open} and Mask-free OVIS~\citep{vs2023mask} suffer from noisy pseudo-masks caused by inaccurate visual-textual alignments, limiting segmentation accuracy. CGG~\citep{wu2023betrayed} addresses this problem by introducing caption grounding and caption generation losses. The grounding loss aligns object nouns in captions with their corresponding image regions while the generation loss enhances contextual understanding.

However, existing open-vocabulary instance segmentation methods still face several fundamental challenges. First, their utilization of captions is typically limited, resulting in insufficient visual-textual grounding. Second, these models struggle to handle synonyms and out-of-vocabulary (OOV) words, thereby constraining their ability to generalize to novel categories and diverse linguistic expressions. Lastly, some approaches rely entirely on noisy pseudo-masks~\citep{vs2023mask}, often leading to suboptimal segmentation accuracy.

To address these challenges, we propose a multimodal framework that leverages pre-trained vision-language models for automatic pseudo-label generation, CLIP-guided synonym filtering, and GPT-based caption reconstruction. In our target-vocabulary-assisted pseudo-labeling setting, target novel-category names are used as text prompts for Grounded SAM during pseudo-mask generation. The framework integrates an automated pipeline built upon Grounded SAM~\citep{groundedsam2024} and LLaVA~\citep{liu2023visual} to generate pseudo segmentation masks, descriptive captions, and semantically aligned synonym sets, thereby enhancing visual-language alignment without requiring any additional manual annotations. The generated pseudo masks, captions, and synonyms are then jointly utilized with the labeled base dataset to train the model. Furthermore, we extend existing classification and mask losses by introducing an enhanced grounding loss, a semantic consistency loss, and a generative caption reconstruction loss. Together, these components enable more robust and semantically coherent visual-textual alignment under the target-vocabulary-assisted protocol, improving generalization across both seen and target novel categories.

Rather than proposing a fundamentally new segmentation architecture, this work focuses on improving open-vocabulary generalization through target-vocabulary-assisted multimodal pseudo-labeling and training-time visual-textual supervision. We adopt a Mask2Former-based segmenter as a strong base architecture and introduce complementary supervision signals generated from pretrained vision-language models. Specifically, Grounded SAM provides pseudo segmentation masks using target novel-category prompts, LLaVA generates descriptive pseudo captions, and CLIP filters visually grounded synonym candidates. These multimodal pseudo-labels are then used with synonym-aware grounding, semantic consistency, and generative caption reconstruction losses to improve robustness to vocabulary variation and unseen categories.

The main contributions of this paper are summarized as follows:
\begin{itemize}
\item We introduce an automated multimodal pipeline that leverages pre-trained vision-language models to generate pseudo segmentation masks, descriptive captions, and semantically aligned synonym sets, providing additional multimodal supervision without manual annotations under a target-vocabulary-assisted pseudo-labeling protocol.
\item We propose a semantic consistency loss and an extended grounding loss that leverage both predefined category names and their visually grounded synonyms to improve generalization and robustness to vocabulary variations.
\item We introduce a GPT-based generative caption reconstruction loss that enhances visual-textual reasoning by reconstructing masked captions conditioned on visual features.
\item We demonstrate that the proposed method consistently outperforms previous state-of-the-art approaches on both OVIS and OSPS benchmarks using the COCO dataset under the target-vocabulary-assisted evaluation protocol.
\end{itemize}

\section{Related Works}
\label{sec:related_works}
\subsection{Open-Vocabulary Instance and Panoptic Segmentation}
Recent surveys have provided comprehensive overviews of open-vocabulary learning and segmentation. \cite{wu2024towards} review open-vocabulary learning in relation to zero-shot learning, open-set recognition, and out-of-distribution detection, and summarize recent progress in open-vocabulary detection and segmentation. Similarly, \cite{zhou2024image} discuss image segmentation in the foundation-model era, highlighting the growing role of large-scale vision-language and promptable segmentation models. Together, these surveys indicate a clear trend from fixed-vocabulary supervised segmentation toward weak image-text supervision, pseudo-label generation, and foundation-model-based visual-textual alignment.

\cite{huynh2022open} introduced one of the earliest frameworks for open-vocabulary instance segmentation, which adopts a teacher-student paradigm. In this framework, the teacher model generates pseudo-masks by aligning the visual features of object regions with the word embeddings of objects extracted from image captions. The student model is then trained using these pseudo-masks while simultaneously estimating the noise levels of the pseudo-mask annotations to improve robustness against imperfect supervision.

Since open-vocabulary methods are typically trained with strong human supervision for base classes and weak supervision from image-caption pairs for novel categories~\citep{huynh2022open}, they tend to exhibit a bias toward base categories. To mitigate this issue, \cite{vs2023mask} proposed the Mask-free OVIS framework, which eliminates the need for manual mask annotations for both base and novel classes. Their approach trains a Mask R-CNN architecture using pseudo-masks generated from a pre-trained vision-language model and image-caption pairs.

However, both XPM~\citep{huynh2022open} and Mask-free OVIS~\citep{vs2023mask} rely on pseudo-masks generated by mapping object regions to words in captions. Consequently, inaccurate visual-textual alignments can produce noisy pseudo-masks, ultimately limiting segmentation accuracy. To address this limitation, \cite{wu2023betrayed} proposed the CGG framework, which introduces a novel caption grounding loss and a caption generation loss. The grounding loss is computed using only object nouns in captions to prevent matching non-visible words to image regions. Meanwhile, the caption generation loss complements the grounding loss by encouraging the model to learn richer contextual representations from image-caption pairs.

Different from previous works~\citep{huynh2022open, vs2023mask, wu2023betrayed}, we propose a unified framework that integrates additional grounding, semantic consistency, and generative caption reconstruction losses based on automatically generated pseudo-captions and CLIP-filtered synonyms, to enhance visual-textual alignment. While CGG~\citep{wu2023betrayed} relies on restricted noun-based grounding, our framework provides richer supervision by automatically generating diverse captions using LLaVA, extracting semantically related words using CLIP, and leveraging them for grounding and semantic consistency objectives. Moreover, by incorporating a GPT-driven caption reconstruction loss, our model captures fine-grained contextual semantics beyond object-level grounding.

\subsection{Mask-Classification and Universal Segmentation Architectures}
Mask-classification-based architectures have become strong foundations for modern segmentation. MaskFormer~\citep{MaskFormer2021} reformulates segmentation as a mask classification problem by predicting a set of binary masks with corresponding class labels, rather than assigning labels independently to each pixel. Mask2Former~\citep{Mask2Former2022} further extends this framework with masked attention and provides a unified architecture for semantic, instance, and panoptic segmentation. Owing to its strong performance and flexibility, Mask2Former~\citep{Mask2Former2022} has been widely adopted as a backbone or baseline architecture in subsequent segmentation studies.

Recent studies have also explored query learning and clustering-based formulations for segmentation. \citet{wang2022learning} introduce instance-unique and transformation-equivariant queries for query-based instance segmentation, while CLUSTSEG~\citep{liang2023clustseg} formulates diverse segmentation tasks as a unified neural clustering process. These works highlight the importance of query-based learning and unified formulations for diverse segmentation tasks.

Recent methods have further explored unified and generalist segmentation frameworks. X-Decoder~\citep{XDecoder2023} introduces a generalized decoding framework that connects pixel-level segmentation outputs and language tokens within a shared semantic space. FreeSeg~\citep{FreeSeg2023} studies universal and open-vocabulary segmentation by integrating dense prediction with flexible language representations. OMG-Seg~\citep{OMGSeg2024} further investigates whether a single model can handle a broad range of segmentation tasks, including image segmentation, video segmentation, open-vocabulary segmentation, prompt-driven segmentation, and interactive segmentation. These methods are relevant to our work because they demonstrate the importance of flexible segmentation architectures for handling diverse categories and tasks.

Our method is related to these architectures because it adopts a Mask2Former-style query-based segmenter. However, unlike methods that introduce a new decoder or segmentation backbone, our work focuses on improving open-vocabulary generalization through multimodal pseudo-labeling and training-time visual-textual supervision.

\subsection{Foundation Models for Open-Vocabulary and Promptable Segmentation}
Foundation models have recently played an important role in open-vocabulary and promptable segmentation. SAM~\citep{kirillov2023segment} provides strong promptable mask generation and demonstrates strong transfer ability across diverse segmentation scenarios. Grounded SAM~\citep{groundedsam2024} combines open-set object localization with mask prediction by integrating Grounding DINO~\citep{liu2023grounding} and SAM~\citep{kirillov2023segment}, enabling the segmentation of arbitrary text-specified objects. Open-Vocabulary SAM~\citep{ovSAM2025} further integrates SAM and CLIP for simultaneous interactive segmentation and recognition over a large vocabulary. ODISE~\citep{ODISE2023} leverages diffusion and vision-language representations for open-vocabulary panoptic segmentation. These works demonstrate the strong potential of foundation models for scalable mask generation, promptable segmentation, and open-vocabulary recognition.

Unlike methods that mainly use foundation models as segmentation or recognition components, our framework uses them primarily to generate complementary supervision signals for training. Specifically, Grounded SAM~\citep{groundedsam2024} provides pseudo segmentation masks, LLaVA~\citep{liu2023visual} generates descriptive pseudo captions, and CLIP~\citep{radford2021learning} filters visually grounded synonym candidates. During inference, these auxiliary modules are removed, and the model retains only the Mask2Former-based segmentation components and CLIP text embeddings.

\begin{table*}[t]
\centering
\caption{Conceptual and technical differences between CGG and the proposed method. Both methods use a Mask2Former-based segmentation architecture, while our method differs in multimodal pseudo-label construction and synonym-aware supervision.}
\label{tab:cgg_comparison}
\small
\begin{tabular}{ >{\centering}m{0.22\textwidth}|  >{\centering}m{0.3\textwidth}  >{\centering\arraybackslash}m{0.4\textwidth} }
\toprule
Component & CGG~\citep{wu2023betrayed} & Ours \\
\midrule
Base architecture & Mask2Former-based & Mask2Former-based \\
Inference-time modules & Segmenter + text embeddings & Segmenter + text embeddings \\
Caption supervision & Original captions & Original captions + LLaVA-generated pseudo captions \\
Grounding target & Object nouns from captions & Object nouns, novel words, and visually grounded synonyms \\
Novel-class pseudo-mask supervision & Not explicitly generated & Grounded-SAM-based pseudo masks \\
Synonym handling & Not explicitly modeled & CLIP-based visual synonym filtering \\
Semantic consistency & No explicit synonym-category consistency & Synonym-category semantic consistency loss \\
Caption objective & Caption generation & GPT-based masked caption reconstruction \\
\bottomrule
\end{tabular}
\end{table*}

\subsection{Language-Supervised Visual-Textual Alignment for Segmentation}
Language-conditioned segmentation methods align visual regions, pixels, or mask embeddings with text representations to recognize categories beyond a fixed vocabulary. OpenSeg~\citep{OpenSeg2022} and related vision-language segmentation approaches exploit image-text supervision to learn open-vocabulary dense prediction. PGSeg~\citep{PGSeg2023} studies weakly open-vocabulary semantic segmentation using only image-text pairs and introduces prototypical knowledge to guide visual grouping and group-text alignment. CLIPSelf~\citep{wu2024clipself} analyzes the gap between global CLIP image representations and local region representations, and adapts CLIP-based vision transformers for open-vocabulary dense prediction without requiring region-text pairs. These methods demonstrate the importance of language supervision and region-language alignment for generalizing to unseen categories.

Our method follows this general direction but differs in two aspects. First, instead of relying only on predefined class names or original captions, we generate pseudo captions and extract visually grounded synonyms to enrich the training vocabulary. Second, we explicitly introduce synonym-aware grounding, semantic consistency, and caption reconstruction losses to strengthen visual-textual alignment at the region and caption levels. This design enables the model to better handle linguistic variations and out-of-vocabulary expressions in open-vocabulary instance and open-set panoptic segmentation.

%\subsection{Other Segmentation Tasks and Applications}
%Recent studies have explored segmentation in various settings, including different supervision types, modalities, and application domains. Segment Any RGB-Thermal Model~\citep{xing2025segment} uses language-aided distillation for RGB-thermal segmentation, and SAM supervision has been applied to 3D weakly supervised point cloud segmentation~\citep{you2025integrating}. EHIN~\citep{Li2026EHIN} studies weakly supervised referring image segmentation, while few-shot image segmentation methods exploit independent query information to improve adaptation to novel classes~\citep{Liu2026Exploiting}. In medical image segmentation, boundary refinement has been studied for colorectal polyp segmentation~\citep{Yue2024Boundary}, and uncertainty-guided backdoors have been used for ownership verification of medical segmentation models~\citep{Yu2026StealthMark}. Physically guided open-vocabulary segmentation incorporates physical priors and weighted patch alignment to improve dense prediction~\citep{Liu2025Physically}.

\begin{figure*}[!t]
\begin{minipage}{1\linewidth}
\centering
\includegraphics[scale=0.6]{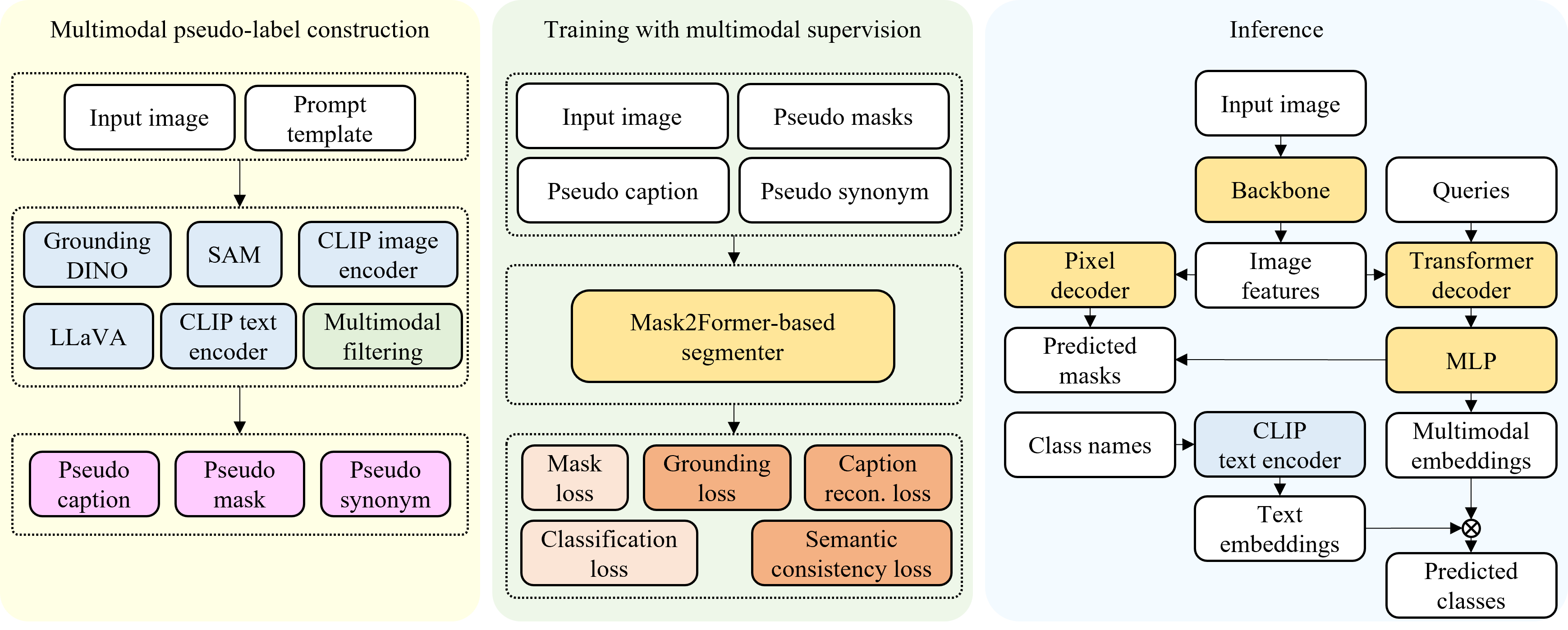}
\end{minipage}
\caption{Overall framework of MCCF. The proposed framework consists of three stages: multimodal pseudo-label construction, training with multimodal supervision, and lightweight inference. First, MCCF constructs pseudo masks, LLaVA-generated pseudo captions, and visually grounded pseudo synonyms. These pseudo labels are then used to train a Mask2Former-based segmenter with standard segmentation losses and the proposed visual-textual objectives, including synonym-aware grounding, semantic consistency, and GPT-based caption reconstruction losses. At inference time, auxiliary pseudo-label generation modules are removed.}
\label{fig:overview}
\end{figure*}

\section{Method}
\label{sec:method}
\subsection{Problem Formulation} 
Following previous works~\citep{huynh2022open, wu2023betrayed}, during training, we are given a training set $\calD_b$ consisting of images $\mI$ and their ground-truth annotations $\calY$ for base classes $\calC_b$ (\ie $\calD_b = \{(\mI_i, \calY_i)\}_{i=1}^{N_b}$) where $N_b$ denotes the number of images in $\calD_b$. Each ground-truth annotation $\calY_i$ includes pairs of an instance mask and its class label for objects belonging to $\calC_b$ in $\mI_i$. In addition, another dataset $\calD_c$ is provided containing images $\mI$ and their image-level captions $C^c$ (\ie $\calD_c = \{(\mI_i, C^c_i)\}_{i=1}^{N_c}$) where $N_c$ represents the number of images in $\calD_c$. We denote the set of object nouns extracted from the captions as caption classes $\calC_c$. Since captions contain a wider variety of words than the limited set of base class names, the number of caption classes $|\calC_c|$ is larger than the number of base classes $|\calC_b|$.

The trained model is expected to predict accurate instance masks and class labels for objects belonging not only to the base classes $\calC_b$ but also to novel categories $\calC_n$. The novel class set $\calC_n$ includes all categories outside $\calC_b$, encompassing both caption classes $\calC_c$ and entirely unseen categories during training. For classification, we employ text embeddings of class names extracted from the pre-trained CLIP text encoder~\citep{radford2021learning}.

\subsection{Baseline Method}
Following CGG~\citep{wu2023betrayed}, we adopt Mask2Former~\citep{Mask2Former2022} as our baseline segmentation architecture because of its high accuracy and ability to handle diverse segmentation tasks including panoptic, instance, and semantic segmentation. As in Mask2Former~\citep{Mask2Former2022}, our baseline architecture consists of three main components: a backbone, a pixel decoder, and a transformer decoder. The pixel decoder generates per-pixel embeddings, while the transformer decoder, followed by an MLP head, produces multimodal embeddings and mask embeddings. Final mask predictions are obtained by computing the product between the per-pixel embeddings and the mask embeddings.

Unlike CGG~\citep{wu2023betrayed}, which employs BERT~\citep{devlin2018bert} for textual embeddings, we use the CLIP text encoder to extract text embeddings, enabling better alignment between visual and textual representations in a shared multimodal space. Class predictions are computed as the dot product between the multimodal embeddings from the transformer decoder and the text embeddings. As illustrated in the inference stage of Figure~\ref{fig:overview}, during inference, we retain only the Mask2Former components~\citep{Mask2Former2022} and the pre-trained text embeddings of all classes, including both base and novel categories. The auxiliary modules used during training are excluded at inference time to ensure efficient prediction.

Table~\ref{tab:cgg_comparison} summarizes the conceptual and technical differences between CGG and the proposed method. While both methods adopt a Mask2Former-based segmentation architecture, they differ in how visual-textual supervision is constructed and exploited during training. Specifically, our method extends CGG by introducing LLaVA-generated pseudo captions, Grounded-SAM-based pseudo masks, CLIP-based synonym filtering, synonym-category embedding alignment, and GPT-based caption reconstruction, while retaining a comparable inference-time structure based on the segmentation model and text embeddings.

\subsection{Proposed Method}
\label{sec:proposed_method}
Figure~\ref{fig:overview} provides an overview of the proposed MCCF framework. The framework is organized into three stages. First, we construct multimodal pseudo labels by combining pseudo masks, LLaVA-generated pseudo captions, and visually grounded pseudo synonyms. Second, these pseudo labels are used to train a Mask2Former-based segmenter with both standard segmentation losses and the proposed visual-textual objectives, including synonym-aware grounding, semantic consistency, and GPT-based caption reconstruction. Finally, during inference, the auxiliary pseudo-label generation modules are removed, and the model performs prediction using only the trained segmenter and CLIP text embeddings. This design enables MCCF to exploit rich multimodal supervision during training while maintaining a lightweight inference pipeline.

\subsubsection{Multimodal Pseudo-label Generation and Refinement}
\label{sec:pseudo_label}
To enhance visual-language alignment without requiring any additional manual annotations, we propose an automated multimodal pipeline that generates pseudo segmentation masks, descriptive captions, and semantically aligned synonym sets using pre-trained vision-language models. The automatically generated pseudo-labels provide both pixel-level and language-level supervision, which are subsequently leveraged during training to improve generalization to unseen categories. As the entire procedure relies only on pre-trained modules and predefined processing steps, it is executed only once before training.

\begin{figure}[!t] 
\begin{minipage}{1\linewidth}
\centering 
\includegraphics[scale=0.6]{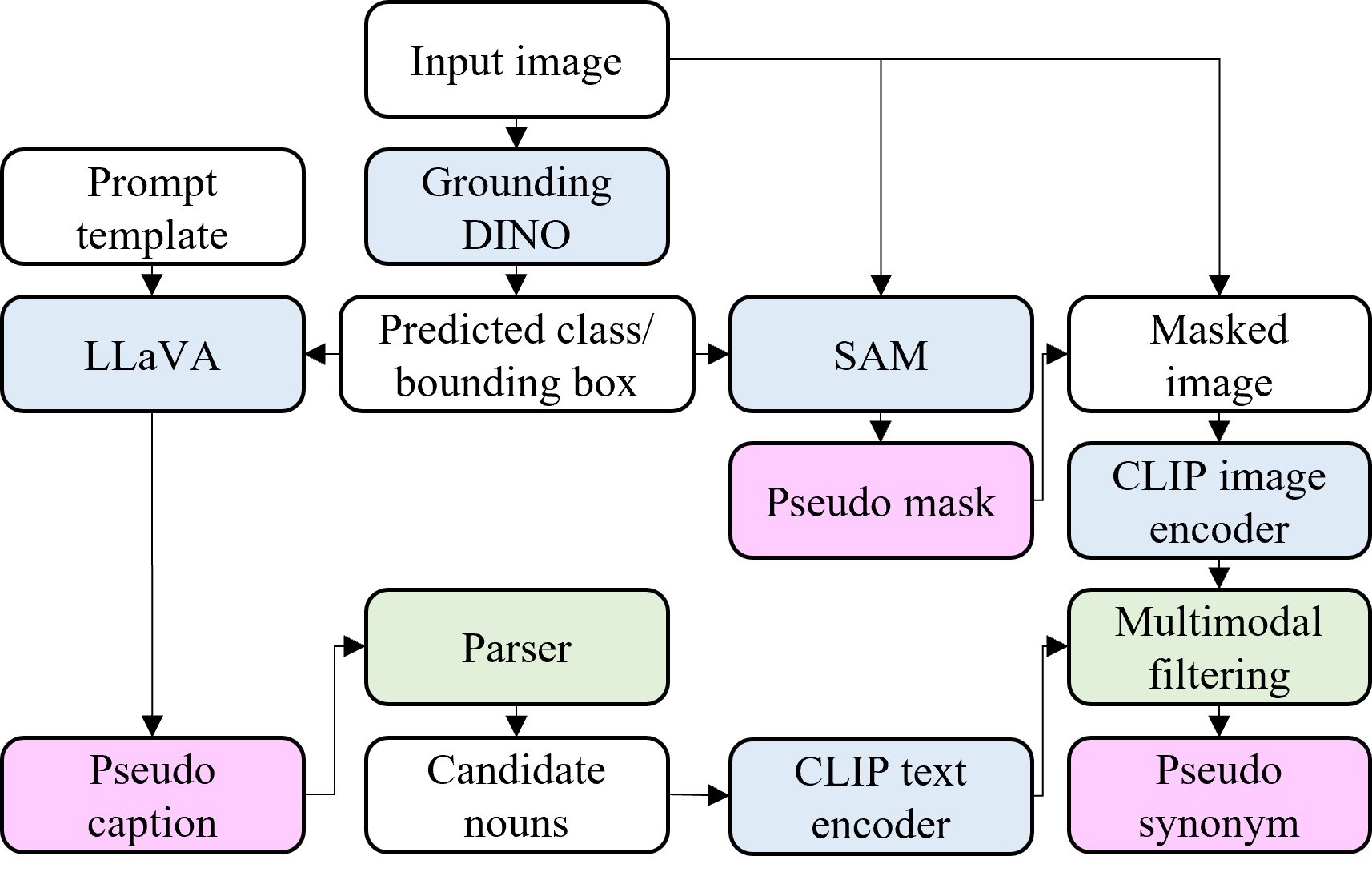}
\end{minipage}
\caption{Multimodal pseudo-label construction. Grounding DINO and SAM generate pseudo masks, LLaVA generates pseudo captions, and CLIP-based multimodal filtering selects visually grounded pseudo synonyms from candidate nouns extracted from the pseudo captions. Blue boxes denote frozen pretrained modules, green boxes denote non-learnable processing steps, pink boxes indicate generated pseudo labels, and white boxes denote inputs or intermediate outputs.}
\label{fig:pseudo_generation}
\end{figure}

\vspace{1mm}
\noindent
\textbf{Pseudo Mask and Caption Generation}.
To generate pseudo masks, we employ Grounded SAM~\citep{groundedsam2024}, which integrates Grounding DINO~\citep{liu2023grounding} for object detection and SAM~\citep{kirillov2023segment} for segmentation, as illustrated in Figure~\ref{fig:pseudo_generation}. Specifically, given an image $\mI$, Grounding DINO detects objects belonging to novel classes $\calC_n$ based on their text labels, producing bounding boxes, confidence scores, and class predictions. These bounding boxes are then refined by SAM into pixel-level pseudo segmentation masks $\mM^{psd}$. As a result, high-quality pseudo-mask annotations are automatically generated for novel categories without any human supervision.

Next, we use LLaVA (Large Language and Vision Assistant)~\citep{liu2023visual} to generate pseudo captions $C^{psd}$ describing the detected objects, as shown in Figure~\ref{fig:pseudo_generation}. LLaVA combines CLIP-ViT~\citep{radford2021learning} for visual encoding and Vicuna~\citep{chiang2023vicuna} for language generation. Given the class labels \textit{\{CLASS\_LABELS\}} predicted by Grounding DINO~\citep{liu2023grounding}, we embed them into a structured prompt designed to encourage LLaVA to produce detailed, diverse, and semantically rich descriptions using synonyms or paraphrased expressions. The prompt is defined as follows:
\begin{quote}
\textit{"There is/are \{CLASS\_LABELS\} in the image. Describe their appearance, position, and quantity in detail and accurately. Instead of using the words in \{CLASS\_LABELS\}, try to use synonyms or creative descriptions to convey their identity."}
\end{quote}

\vspace{1mm}
\noindent
\textbf{Multimodal Filtering-based Pseudo Synonym Generation}.
Given the pseudo captions generated by LLaVA, we first extract nouns and noun phrases and identify candidate words that do not belong to the pre-defined class vocabulary, as illustrated in Figure~\ref{fig:pseudo_generation}. Then, we propose a CLIP-based multimodal filtering strategy to select visually grounded synonym candidates among these undefined words.

Specifically, for each undefined word $w^{undef}$ extracted from the pseudo captions, we encode it using the CLIP text encoder to obtain its textual embedding $\vt$. For the visual representation, we crop and mask the input image $\mI$ using the pseudo masks $\mM^{psd}$ generated by SAM for novel categories. The masked image is computed as $\mI^{mask} = \mI \odot \mM^{psd}$, where $\odot$ denotes element-wise multiplication. These masked images are then encoded using the CLIP image encoder to obtain visual embeddings $\vv$.

Then, we compute the multimodal similarity between the textual embedding $\vt_i^{undef}$ of the $i$-th word and the visual embedding $\vv_j^{mask}$ of the $j$-th masked image using cosine similarity:
\begin{equation}
s_{ij} = \frac{(\vt_i^{undef})^T \vv_j^{mask}}{\norm{\vt_i^{undef}} \norm{\vv_j^{mask}}},
\end{equation}
where $\norm{\cdot}$ denotes the Euclidean norm. The similarity score quantifies the semantic and visual alignment between textual concepts and localized visual regions.

To retain only the most relevant and visually consistent synonym candidates, we apply a filtering strategy that combines top-1 ranking with a similarity threshold. A candidate synonym $w^{syn}$ is preserved if its similarity score exceeds a threshold of $\tau = 0.4$ and ranks as the most similar word for the corresponding visual instance. This criterion suppresses noisy or ambiguous terms while retaining synonyms that are semantically and visually consistent with masked images for novel categories.

The threshold $\tau=0.4$ is empirically selected to balance the retention of visually relevant synonym candidates and the suppression of noisy or weakly grounded words. We use top-1 selection together with this threshold to keep only the most visually consistent synonym for each pseudo instance.

\subsubsection{Training}
\begin{figure*}[!t] 
\begin{minipage}{1\linewidth}
\centering 
\includegraphics[scale=0.6]{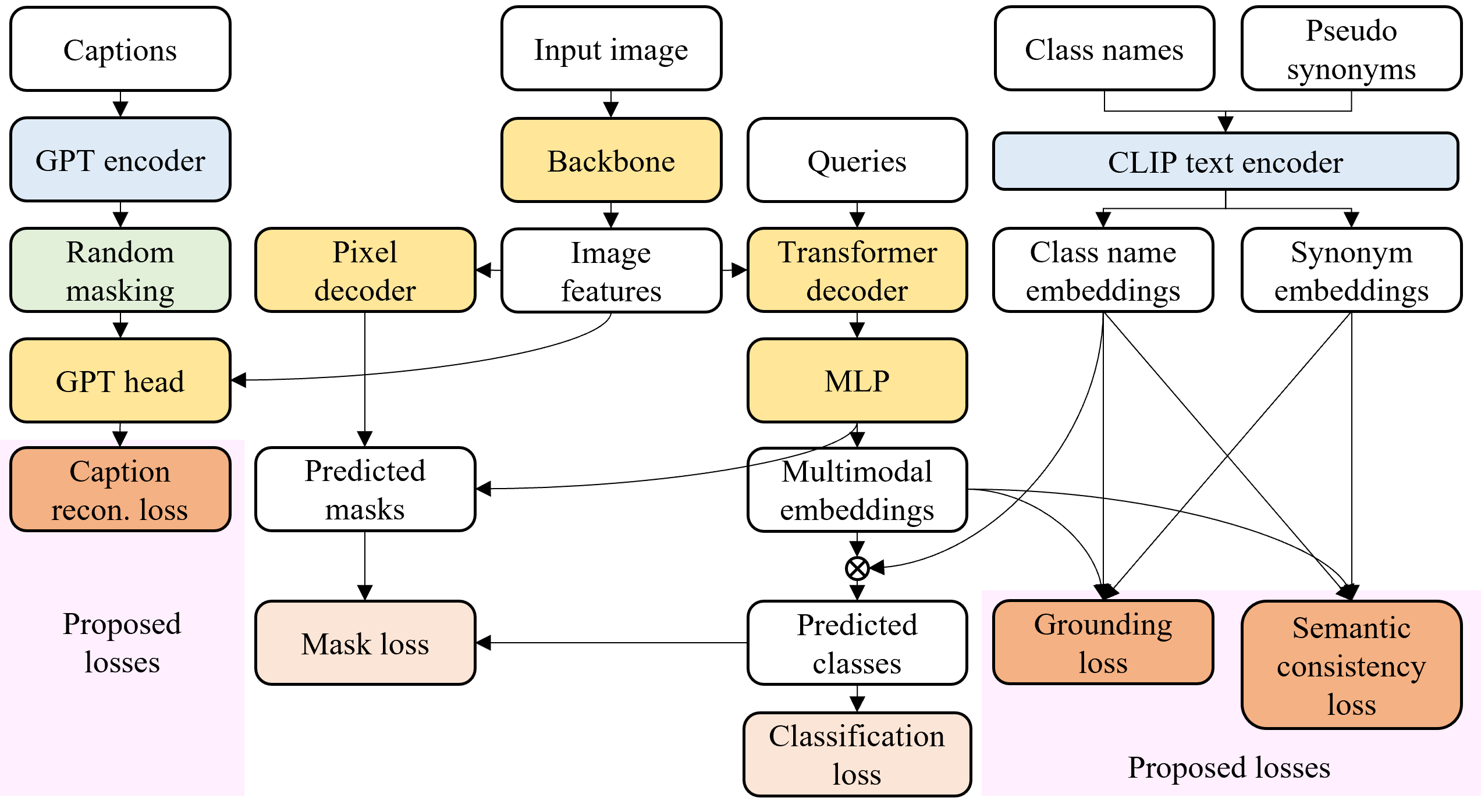}
\end{minipage}
\caption{Training framework of MCCF. The Mask2Former-based segmenter is trained with pseudo masks, class names, pseudo synonyms, and captions. In addition to standard classification and mask losses, MCCF introduces synonym-aware grounding, semantic consistency, and GPT-based caption reconstruction losses. Blue boxes denote frozen pretrained modules, yellow boxes denote trainable network components, green boxes denote non-learnable processing steps, orange boxes denote loss functions, and white boxes denote inputs or intermediate outputs.}
\label{fig:framework_training}
\end{figure*}

We leverage the pseudo segmentation masks $\mM^{psd}$, pseudo captions $C^{psd}$, and pseudo synonyms $w^{psd}$ generated for the dataset $\calD_c$ from~\sref{sec:pseudo_label}, in addition to the labeled training set $\calD_b$ for base classes, during training. We adopt the classification and mask losses from~\citep{wu2023betrayed, Mask2Former2022}, enhance the grounding loss from~\citep{wu2023betrayed} using the pseudo synonyms, and further introduce a semantic consistency loss and a generative caption reconstruction loss. The overall training framework is illustrated in Figure~\ref{fig:framework_training}.

\vspace{1mm}
\noindent
\textbf{Classification Loss}. 
We employ a cross-entropy loss for the classification loss $\calL_{\text{cls}}$ to align visual and textual embeddings using $\calD_b$ and $\calD_c$ with pseudo labels. The logits are obtained from the dot product between the multimodal embeddings $\vf$ produced by the transformer decoder (with an MLP head) and the text embeddings $\vt$ extracted from the CLIP text encoder, as shown in Figure~\ref{fig:framework_training}. Formally, the classification loss $\calL_{\text{cls}}$ is defined as:
\begin{equation}
\calL_{\text{cls}} = - \sum_{i} \vy_i \ln(\vp_i),
\end{equation}
where $i$ indexes the classes, $\vy$ denotes the one-hot encoded class label of an instance, and $\vp_i$ represents the predicted probability for class $i$, obtained by applying a softmax to the dot product between $\vf$ and $\vt$.

\vspace{1mm}
\noindent
\textbf{Mask Loss}. 
We adopt the mask loss $\calL_{\text{mask}}$ from~\citep{Mask2Former2022}, which comprises a mask classification loss $\calL_{\text{mask-cls}}$, a pixel-wise binary cross-entropy loss $\calL_{\text{ce}}$, and a Dice loss $\calL_{\text{dice}}$. The total mask loss $\calL_{\text{mask}}$ is computed using $\calD_b$ and $\calD_c$ with pseudo labels and is formulated as:
\begin{equation}
\calL_{\text{mask}} = \lambda_{\text{mask-cls}} \calL_{\text{mask-cls}} + \lambda_{\text{ce}} \calL_{\text{ce}} + \lambda_{\text{dice}} \calL_{\text{dice}},
\end{equation}
where $\lambda_{\text{mask-cls}}$, $\lambda_{\text{ce}}$, and $\lambda_{\text{dice}}$ are weighting coefficients that balance the contribution of each term.

\vspace{1mm}
\noindent
\textbf{Grounding Loss}. 
We adopt the grounding loss from~\citep{wu2023betrayed}, which aims to learn visual-textual alignment by maximizing the similarity between the visual and textual embeddings of matched pairs. However, while CGG~\citep{wu2023betrayed} computes this loss using only object nouns extracted from the given captions in $\calD_c$, we extend it to include both object nouns and their synonyms obtained from the given captions $C^c$ and the generated pseudo captions $C^{psd}$. This extension enables the model to capture a broader range of semantic relationships between visual regions and language expressions.

Specifically, we compute cosine similarities between the visual embeddings $\vf$ from the transformer decoder (with an MLP head) and the text embeddings $\vt^{nov}$ and $\vt^{syn}$ for novel class names and their synonyms, respectively, using the CLIP text encoder. Then, the grounding loss is designed to maximize similarity for matched visual–text pairs while minimizing it for mismatched ones. The overall grounding loss $\calL_{\text{gr}}$ is defined as follows:
\begin{equation}
\calL_{\text{gr}} = \calL_{\text{gr-nov}}(\vf, \vt^{novel}) + \calL_{\text{gr-syn}}(\vf, \vt^{syn}),
\label{loss:grounding}
\end{equation}
where $\calL_{\text{gr-nov}}$ and $\calL_{\text{gr-syn}}$ denote the grounding losses for novel-category nouns and their synonyms, respectively. By optimizing this objective, the model learns a unified visual-text embedding space that effectively associates image regions with diverse textual concepts, thereby enhancing zero-shot recognition and robustness to vocabulary variations.

\vspace{1mm}
\noindent
\textbf{Semantic Consistency Loss}. 
This loss aims to enforce consistency among semantically equivalent words, particularly novel class names and their synonyms, within a shared multimodal embedding space. Although synonyms such as “plane,” “airplane,” and “aircraft” convey the same meaning, their embeddings may occupy different positions in the feature space, leading to inconsistent predictions.

To mitigate this issue, we project novel class names $w^{nov}$ and their corresponding synonyms $w^{syn}$ into the text embedding space using the CLIP text encoder, yielding $\vt^{nov}$ and $\vt^{syn}$, respectively.  We then compute the distance between these paired embeddings, weighted by the visual embedding $\vf$ obtained from the transformer decoder to emphasize visually relevant regions. The semantic consistency loss $\calL_{\text{cons}}$ is defined as follows:
\begin{equation}
\calL_{\text{cons}} = \{\vf_i^T (\vt^{nov}_i - \vt^{syn}_i)\}^2.
\end{equation}
This loss encourages consistent representations for words with equivalent meanings, enhancing robustness to linguistic variation and improving generalization to unseen categories.

\vspace{1mm}
\noindent
\textbf{Generative Caption Reconstruction Loss}. 
This loss aims to enhance the model's ability to understand and reason over fine-grained visual-textual relationships by incorporating a generative language component. Unlike the grounding loss, which focuses on maximizing the similarity between matched visual and textual embeddings, this loss encourages the model to reconstruct complete captions conditioned on visual features.

Given a caption $C = [c_1, c_2, \dots, c_n]$ from either the original captions $C^c$ or the generated pseudo captions $C^{psd}$, a subset of tokens is randomly replaced with the mask token \texttt{[MASK]}, resulting in a masked caption $\tilde{C}$. The masked caption $\tilde{C}$ is then fed into a GPT-based generative module as the query sequence, while the image features $\mF$ extracted from the backbone of the segmentation network are used as keys and values in the cross-attention layers. This design enables the GPT module to leverage visual information when reconstructing the missing textual tokens.

The generative caption reconstruction loss $\calL_{\text{recon}}$ is computed by maximizing the likelihood of correctly predicting each masked token conditioned on the preceding tokens, the masked caption, and the corresponding image features. Formally, the loss is defined as:
\begin{equation}
\calL_{\text{recon}} = - \sum_{i=1}^{n} \log p(\hat{c}_i \mid c_{i-1}, \tilde{C}, \mF),
\end{equation}
where $p(\hat{c}_i \mid c_{i-1}, \tilde{C}, \mF)$ denotes the conditional probability of predicting token $\hat{c}_i$ given the previous token $c_{i-1}$, the partially masked caption $\tilde{C}$, and the image feature map $\mF$.

\vspace{1mm}
\noindent
\textbf{Total Loss}. 
The total loss $\calL_{\text{total}}$ is defined as the weighted sum of the five previously introduced components: the classification loss $\calL_{\text{cls}}$, mask loss $\calL_{\text{mask}}$, grounding loss $\calL_{\text{gr}}$, semantic consistency loss $\calL_{\text{cons}}$, and generative caption reconstruction loss $\calL_{\text{recon}}$.
\begin{equation}
\calL_{\text{total}} = \calL_{\text{cons}} + \lambda_{\text{cls}} \calL_{\text{cls}} + \lambda_{\text{mask}} \calL_{\text{mask}} + \lambda_{\text{gr}} \calL_{\text{gr}} + \lambda_{\text{recon}} \calL_{\text{recon}},
\label{eq:loss_total}
\end{equation}
where $\lambda_{\text{cls}}$, $\lambda_{\text{mask}}$, $\lambda_{\text{gr}}$, and $\lambda_{\text{recon}}$ are hyperparameters that balance the contribution of each loss term. In all experiments, we set $\lambda_{\text{cls}}$, $\lambda_{\text{mask}}$, $\lambda_{\text{gr}}$, and $\lambda_{\text{recon}}$ to 2, 5, 2, and 2, respectively.

\section{Experiments and Results}
\label{sec:result}
\subsection{Experimental Setting}
\noindent \textbf{Datasets}.
We conduct experiments on the COCO dataset~\citep{lin2014microsoft} for both open-vocabulary instance segmentation (OVIS) and open-set panoptic segmentation (OSPS). For OVIS, following previous works~\citep{zheng2021zero, huynh2022open, wu2023betrayed, vs2023mask}, we split the 65 object classes into 48 base classes and 17 novel classes, where pixel-level annotations are provided only for the base classes during training. The 17 novel categories are bus, dog, cow, elephant, umbrella, tie, skateboard, cup, knife, cake, couch, keyboard, sink, scissors, airplane, cat, and snowboard.

For OSPS, following~\citep{hwang2021exemplar, xu2022two, wu2023betrayed}, we construct three unknown-category settings with 5\%, 10\%, and 20\% unknown thing classes. In the 5\% setting, car, cow, pizza, and toilet are treated as unknown classes. In the 10\% setting, boat, tie, zebra, and stop sign are additionally excluded. In the 20\% setting, dining table, banana, bicycle, cake, sink, cat, keyboard, and bear are further excluded. These categories are removed from the labeled training set and evaluated as unknown thing classes.

\vspace{1mm}
\noindent
\textbf{Metric and Evaluation Protocol}. 
For OVIS, we use the mask-based mean Average Precision (mAP) at an IoU threshold of 0.5. Following~\citep{zareian2021open, huynh2022open, wu2023betrayed, vs2023mask}, we evaluate under two settings: constrained and generalized. In the constrained setting, the trained model is tested on images containing either base or novel classes. In the generalized setting, evaluation is conducted on images containing both base and novel classes. The generalized setting is more challenging due to inherent class bias, which tends to favor the base classes.

For OSPS, we report Panoptic Quality (PQ) and Segmentation Quality (SQ), following~\citep{wu2023betrayed}.

\begin{table*}[t]
\centering
\caption{Quantitative comparison of open-vocabulary instance segmentation. Bold and underlined values denote the best and second-best scores, respectively. $^\dagger$ indicates CGG re-trained using the same Grounded-SAM-based novel-class pseudo-mask annotations as our method while keeping the original CGG training losses.}
\label{tab:ovis}
\small
\begin{tabular}{ >{\centering}m{0.4\textwidth}| >{\centering}m{0.07\textwidth} >{\centering}m{0.07\textwidth}| >{\centering}m{0.07\textwidth}  >{\centering\arraybackslash}m{0.07\textwidth} }
\toprule
\multirow{2}{*}{Method} & \multicolumn{2}{c|}{Constrained} & \multicolumn{2}{c}{Generalized} \\
\cmidrule(lr){2-3} \cmidrule(lr){4-5}
& Base & Novel & Base & Novel \\
\midrule
OVR+OMP~\citep{biertimpel2021prior} & 31.3 & 14.1 & 30.5 & 8.3 \\
SB~\citep{bansal2018zero} & 41.6 & 20.8 & 41.0 & 16.0 \\
BA-RPN~\citep{zheng2021zero} & 41.8 & 20.1 & 41.3 & 15.4 \\
Soft-Teacher~\citep{xu2021end} & 41.8 & 14.8 & 41.5 & 9.6 \\
Unbiased-Teacher~\citep{liu2021unbiased} & 41.8 & 15.1 & 41.4 & 9.8 \\
OVR-RCNN~\citep{zareian2021open} & 42.0 & 20.9 & 41.6 & 17.1 \\
XPM~\citep{huynh2022open} & 42.4 & 24.0 & 41.5 & 21.6 \\
Mask-free OVIS~\citep{vs2023mask} & 36.7 & 27.4 & 28.7 & 25.0 \\
CGG~\citep{wu2023betrayed} & 46.8 & 29.5 & 46.0 & 28.4 \\
\midrule
CGG$^\dagger$ & \textbf{48.0} & \underline{45.1} & \textbf{47.7} & \underline{43.6} \\
MCCF (Ours) & \underline{47.8} & \textbf{51.6} & \underline{47.4} & \textbf{50.4} \\
\bottomrule
\end{tabular}
\end{table*}

\vspace{1mm}
\noindent
\textbf{Implementation Details}. 
We adopt CLIP embeddings~\citep{radford2021learning} as the shared representation space for the classification head, image encoder, and text encoder. For text processing, we employ the BPE tokenizer~\citep{sennrich2015neural} from GPT-2 to tokenize captions and map discrete tokens to continuous embeddings using GPT-2's pre-trained embedding layer, following~\citep{zhu2022exploring}. Unlike~\citep{wu2023betrayed}, which uses an LVIS-specific class parser to extract object nouns, we adopt the NLTK parser for word extraction to preserve lexical diversity and capture a richer semantic vocabulary in open-vocabulary settings.

For OVIS, we retain the top 100 queries as model outputs, corresponding to the highest-confidence object predictions. For OSPS, we follow~\citep{hwang2021exemplar, xu2022two, wu2023betrayed} by prioritizing mask predictions for thing classes and assigning the remaining background regions to stuff categories. The models are trained on two GPUs with a mini-batch size of four using the AdamW optimizer~\citep{loshchilov2017decoupled} with a weight decay of 0.0001. Following~\citep{Mask2Former2022, wu2023betrayed}, we apply random cropping during both pre-training and training.

Our training procedure consists of two stages, following~\citep{wu2023betrayed, zareian2021open}. In the first stage, we perform class-agnostic pretraining using $\calD_b$ and $\calD_c$ with pseudo mask labels without incorporating captions. In the second stage, the model is fine-tuned with all proposed loss functions using $\calD_b$ and $\calD_c$ along with the complete set of pseudo labels.

For the GPT-based caption reconstruction module, we use captions from both the original COCO captions and the LLaVA-generated pseudo captions. During training, a subset of caption tokens is randomly replaced with the [MASK] token. The masked caption embeddings are fed into the GPT-based decoder as the query sequence, while image features extracted from the segmentation backbone are used as keys and values in the cross-attention layers. The decoder then reconstructs the masked tokens conditioned on visual features. This reconstruction objective is used only during training and is optimized in the second training stage together with the segmentation, grounding, and semantic consistency losses.

For pseudo-label noise handling, we use CLIP-based multimodal filtering to remove weakly grounded synonym candidates. Specifically, each candidate synonym extracted from LLaVA-generated pseudo captions is compared with the corresponding masked visual region using CLIP similarity. A candidate synonym is retained only when its similarity score exceeds $\tau=0.4$ and ranks within the top-1 candidates for the corresponding visual instance. This strategy suppresses ambiguous or visually inconsistent words while preserving synonyms that are semantically and visually aligned with the pseudo masks.

\vspace{1mm}
\noindent
\textbf{Pseudo-label vocabulary}.
In our main pseudo-labeling setting, we use the target novel-category vocabulary as text prompts for Grounding DINO~\citep{liu2023grounding} to generate pseudo annotations for novel categories. Therefore, this setting should be interpreted as a target-vocabulary-assisted pseudo-labeling protocol rather than a fully category-agnostic open-vocabulary protocol. Since this protocol can provide stronger pseudo masks than caption-only pseudo-labeling, we additionally introduce a controlled baseline, CGG$^\dagger$, to separate the effect of pseudo-mask quality from the effect of the proposed training objectives. CGG$^\dagger$ is trained with the same Grounded-SAM-based novel-class pseudo-mask annotations as ours while retaining the original CGG loss functions~\citep{wu2023betrayed}. Thus, the comparison with CGG$^\dagger$ provides a more controlled evaluation of the proposed synonym-aware grounding, semantic consistency, and caption reconstruction objectives.

\begin{table*}[!t]
\centering
\caption{Quantitative comparison of open-set panoptic segmentation. The "Unk." column denotes the proportion of unknown classes among all classes. Bold and underlined values indicate the best and second-best scores, respectively. $^\ast$ indicates that the scores are averaged across unknown classes, whereas EOPSN~\citep{hwang2021exemplar} and Dual~\citep{xu2022two} treat all unknown categories as a single class. The superscripts "Th" and "St" denote thing and stuff classes, respectively.}
\label{tab:osps}
\small
\begin{tabular}{>{\centering}m{0.3\textwidth}| >{\centering}m{0.04\textwidth}|>{\centering}m{0.06\textwidth} >{\centering}m{0.06\textwidth}| >{\centering}m{0.06\textwidth} >{\centering}m{0.06\textwidth}| >{\centering}m{0.06\textwidth}  >{\centering\arraybackslash}m{0.06\textwidth}}
\toprule
\multirow{2}{*}{Method} & Unk. & \multicolumn{4}{c|}{Known} & \multicolumn{2}{c}{Unknown} \\
\cmidrule{3-8}
     &  (\%) & $\text{PQ}^{\text{Th}}$ & $\text{SQ}^{\text{Th}}$ & $\text{PQ}^{\text{St}}$ & $\text{SQ}^{\text{St}}$ &$\text{PQ}^{\text{Th}}$ & $\text{SQ}^{\text{Th}}$ \\
\midrule
EOPSN~\citep{hwang2021exemplar} & \multirow{4}{*}{20} & 45.0 & 80.3 & 28.2 & 71.2 & 11.3 & 73.8 \\
Dual~\citep{xu2022two} & & 45.0 & 80.6 & 28.1 & 70.1 & 21.4 & 79.1 \\
CGG~\citep{wu2023betrayed} & & \textbf{48.4} & \textbf{82.3} & \textbf{34.4} & \textbf{81.1} & \underline{36.5}$^\ast$ & \underline{78.0}$^\ast$ \\
MCCF (ours) & & \underline{47.1} & \underline{81.5} & \underline{33.9} & \underline{80.2} & \textbf{54.5}$^\ast$ & \textbf{85.3}$^\ast$ \\        
\midrule
EOPSN~\citep{hwang2021exemplar} & \multirow{4}{*}{10} & 44.5 & 80.6 & 28.4 & 71.8 & 17.9 & 76.8 \\
Dual~\citep{xu2022two} & & 45.0 & 80.7 & 27.8 & 72.2 & 24.5 & 79.9 \\
CGG~\citep{wu2023betrayed} & & \textbf{49.2} & \textbf{82.8} & \textbf{34.6} & \textbf{81.2} & \underline{41.6}$^\ast$ & \underline{82.6}$^\ast$ \\
MCCF (ours) & & \underline{48.4} & \underline{82.3} & \underline{34.0} & \underline{80.8} & \textbf{53.1}$^\ast$ & \textbf{83.7}$^\ast$ \\
\midrule
EOPSN~\citep{hwang2021exemplar} & \multirow{4}{*}{5} & 44.8 & 80.5 & 28.3 & 73.1 & 23.1 & 74.7 \\
Dual~\citep{xu2022two} &  & 45.1 & 80.9 & 28.1 & 73.1 & 30.2 & 80.0 \\
CGG~\citep{wu2023betrayed} & & \textbf{50.2} & \textbf{83.1} & \textbf{34.3} & \textbf{81.5} & \underline{45.0}$^\ast$ & \textbf{85.2}$^\ast$ \\
MCCF (ours) & & \underline{49.8} & \underline{82.9} & \underline{33.8} & \underline{81.3} & \textbf{52.3}$^\ast$ & \underline{83.0}$^\ast$ \\
\bottomrule
\end{tabular}
\end{table*}

\subsection{Results}
\noindent 
\textbf{Quantitative Results for OVIS}. 
Table~\ref{tab:ovis} presents quantitative comparisons between the proposed method and prior works~\citep{wu2023betrayed, vs2023mask, huynh2022open, zareian2021open, liu2021unbiased, xu2021end, zheng2021zero, bansal2018zero, biertimpel2021prior} on the open-vocabulary instance segmentation task. Compared with the original CGG~\citep{wu2023betrayed}, our method improves novel-class AP by 22.1 and 22.0 points under the constrained and generalized settings, respectively. For the base classes, our method achieves improvements of 1.0 and 1.4 AP points in the constrained and generalized settings, respectively.

To further isolate the effect of the proposed supervision strategy, we additionally report CGG$^\dagger$, where CGG is re-trained using the same Grounded-SAM-based novel-class pseudo-mask annotations as our method while retaining the original CGG losses. Compared with CGG$^\dagger$, MCCF improves novel-class AP by 6.5 and 6.8 points in the constrained and generalized settings, respectively. These results indicate that the proposed synonym-aware grounding, semantic consistency, and caption reconstruction objectives provide additional gains beyond stronger pseudo-mask annotations. A detailed ablation study on each proposed objective is further provided in Table~\ref{tab:tbl_ablation_ovis}.

\begin{figure}[!t]
\centering
\begin{minipage}{0.48\linewidth}
\centering 
\includegraphics[width=1\textwidth,height=0.1\textheight]{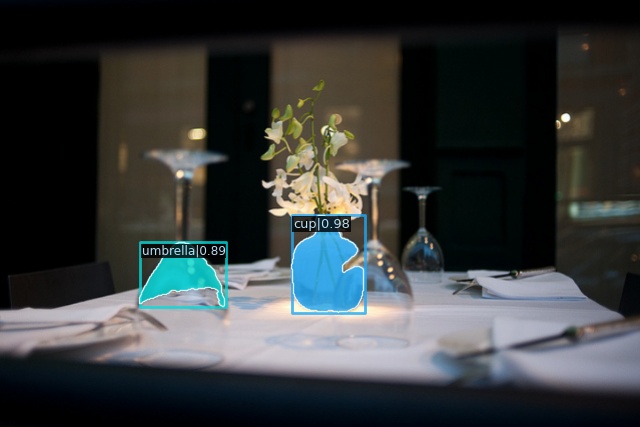}
\end{minipage}
\begin{minipage}{0.48\linewidth}
\centering 
\includegraphics[width=1\textwidth,height=0.1\textheight]{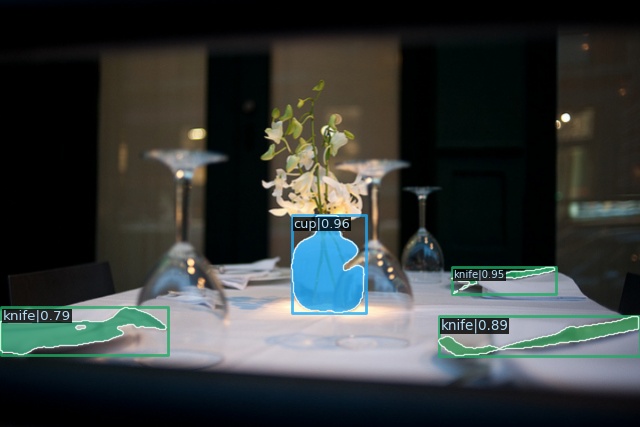}
\end{minipage}

\begin{minipage}{0.48\linewidth}
\centering 
\includegraphics[width=1\textwidth,height=0.1\textheight]{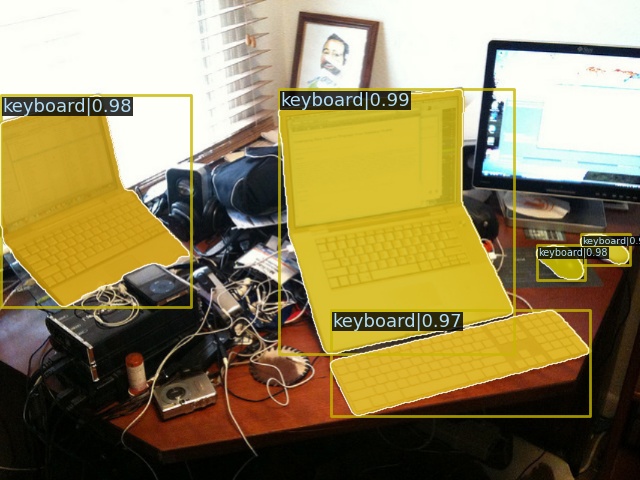}
\end{minipage}
\begin{minipage}{0.48\linewidth}
\centering 
\includegraphics[width=1\textwidth,height=0.1\textheight]{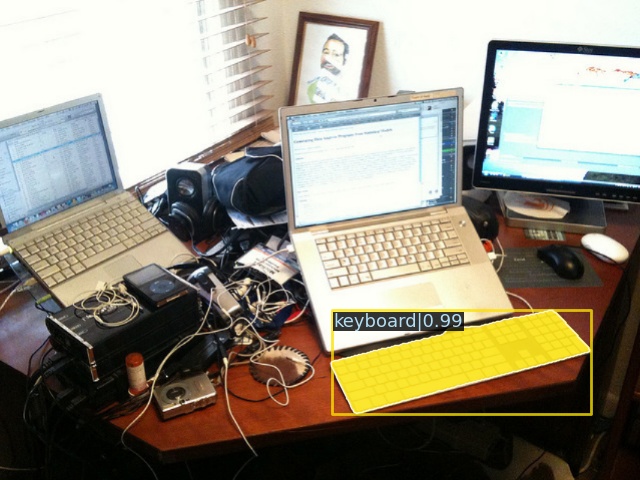}
\end{minipage}

\begin{minipage}{0.48\linewidth}
\centering 
\includegraphics[width=1\textwidth,height=0.1\textheight]{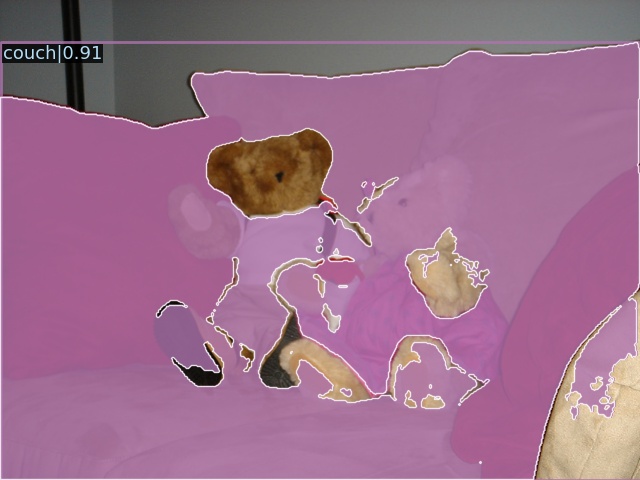}
\end{minipage}
\begin{minipage}{0.48\linewidth}
\centering 
\includegraphics[width=1\textwidth,height=0.1\textheight]{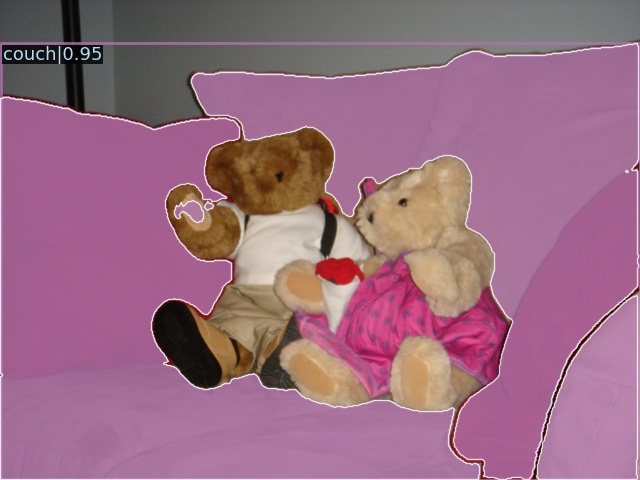}
\end{minipage}

\begin{minipage}{0.48\linewidth}
\centering 
\includegraphics[width=1\textwidth,height=0.1\textheight]{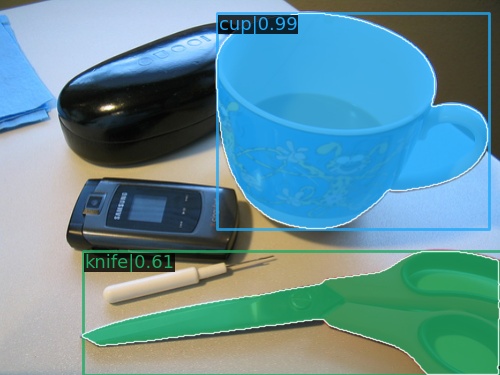}
\end{minipage}
\begin{minipage}{0.48\linewidth}
\centering 
\includegraphics[width=1\textwidth,height=0.1\textheight]{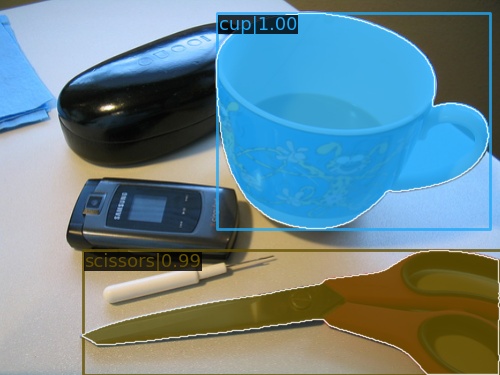}
\end{minipage}

\begin{minipage}{0.48\linewidth}
\centering 
\includegraphics[width=1\textwidth,height=0.1\textheight]{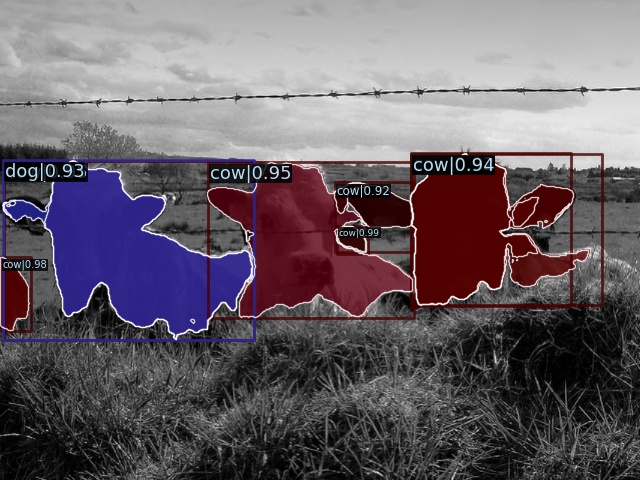}
\end{minipage}
\begin{minipage}{0.48\linewidth}
\centering 
\includegraphics[width=1\textwidth,height=0.1\textheight]{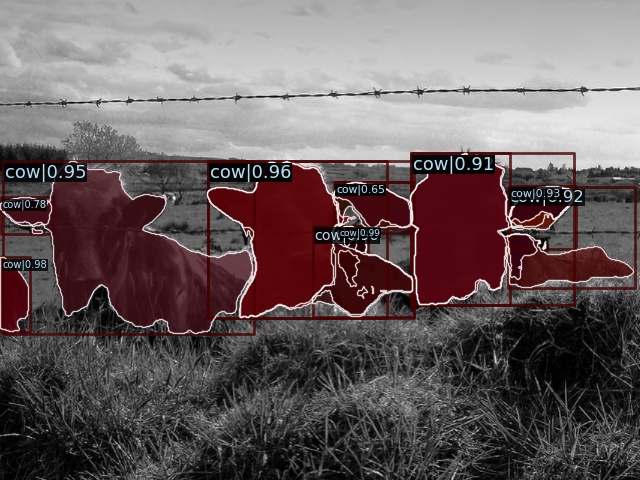}
\end{minipage}
\\
\vspace{1mm}

\begin{minipage}{0.48\linewidth}
\centering 
\small
{(a) CGG}
\end{minipage}
\begin{minipage}{0.48\linewidth}
\centering 
\small
{(b) Ours}
\end{minipage}
\caption{Qualitative comparison of open-vocabulary instance segmentation results between the proposed method and the previous state-of-the-art method, CGG~\citep{wu2023betrayed}.}
\label{fig:result_ovis}
\end{figure}

\vspace{1mm}
\noindent
\textbf{Quantitative Results for OSPS}.
Table~\ref{tab:osps} presents quantitative comparisons between the proposed method and previous works~\citep{wu2023betrayed, xu2022two, hwang2021exemplar} on the open-set panoptic segmentation task. The results show that our method improves unknown-class panoptic segmentation performance over the previous state-of-the-art method~\citep{wu2023betrayed} across most settings and metrics. Specifically, our method achieves absolute improvements of 18.0, 11.5, and 7.3 PQ points for unknown classes under the 20\%, 10\%, and 5\% unknown settings, respectively.

For known classes, our method slightly underperforms CGG~\citep{wu2023betrayed}. This suggests that the proposed training strategy shifts the model toward better recognition of unknown categories, with a small trade-off on known categories. Nevertheless, the substantial gains on unknown classes indicate that the proposed multimodal pseudo-labeling and visual-textual alignment objectives are effective for open-set panoptic segmentation.

\begin{figure*}[!t]
\centering
\begin{minipage}{0.25\linewidth}
\centering 
\includegraphics[width=1\textwidth,height=0.12\textheight]{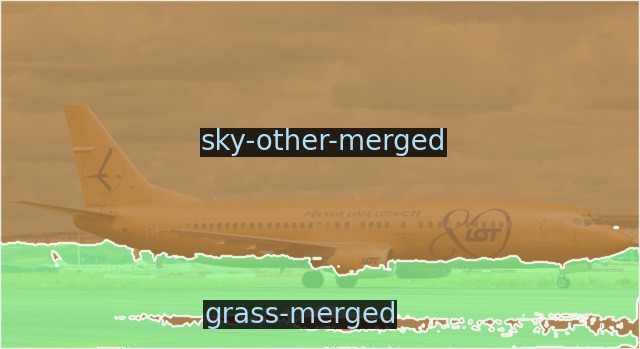}
\end{minipage}
\begin{minipage}{0.25\linewidth}
\centering 
\includegraphics[width=1\textwidth,height=0.12\textheight]{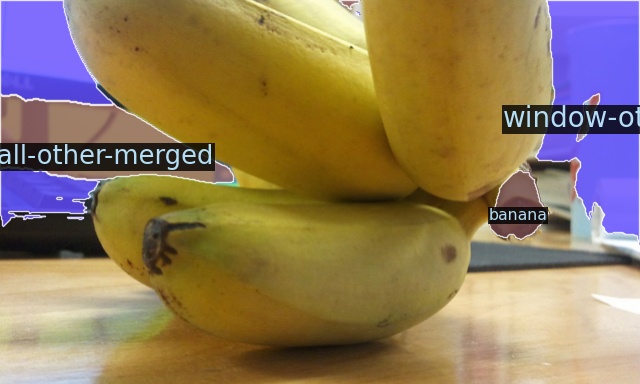}
\end{minipage}
\begin{minipage}{0.25\linewidth}
\centering 
\includegraphics[width=1\textwidth,height=0.12\textheight]{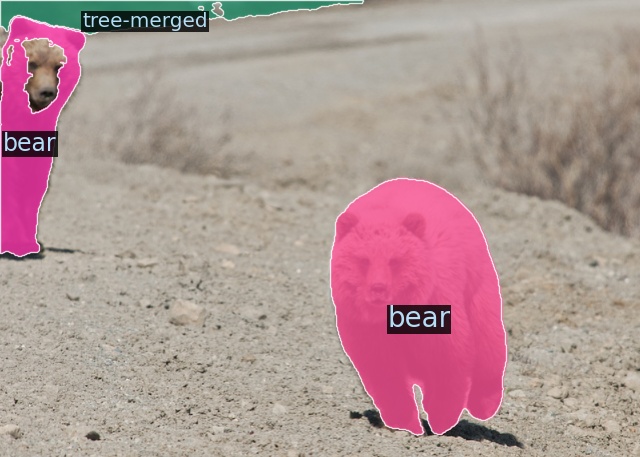}
\end{minipage}

\begin{minipage}{0.25\linewidth}
\centering 
\includegraphics[width=1\textwidth,height=0.12\textheight]{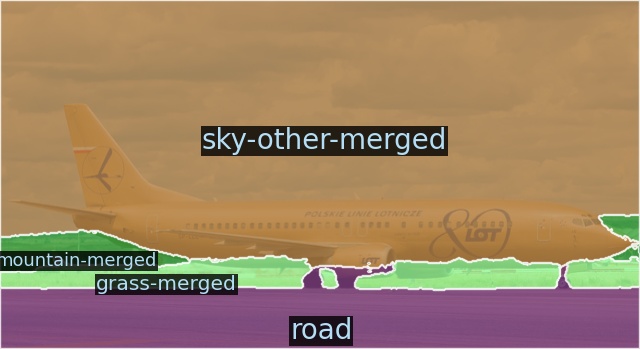}
\end{minipage}
\begin{minipage}{0.25\linewidth}
\centering 
\includegraphics[width=1\textwidth,height=0.12\textheight]{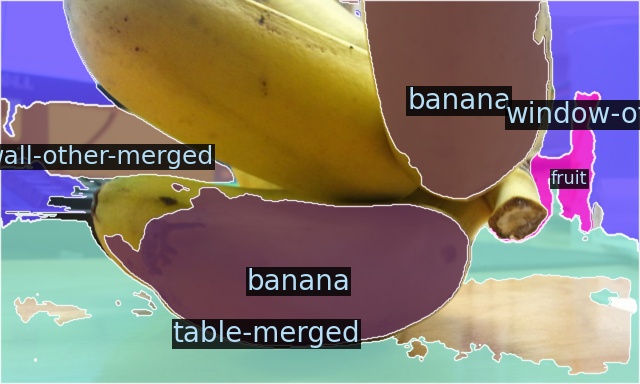}
\end{minipage}
\begin{minipage}{0.25\linewidth}
\centering 
\includegraphics[width=1\textwidth,height=0.12\textheight]{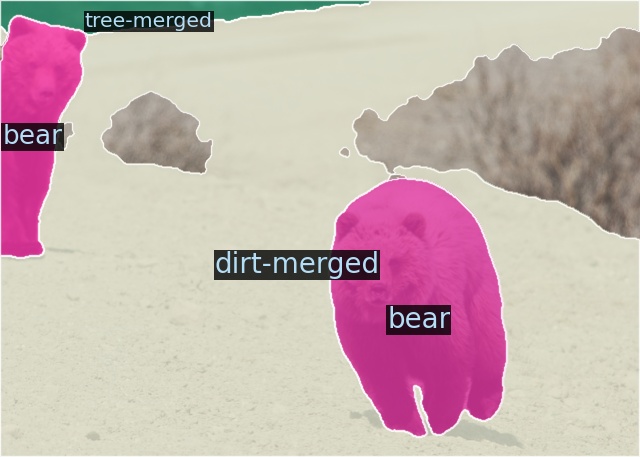}
\end{minipage}
\caption{Qualitative comparison of open-set panoptic segmentation results between the proposed method (second row) and the previous state-of-the-art method, CGG~\citep{wu2023betrayed} (first row).}
\label{fig:result_osps}
\end{figure*}

\begin{figure*}[!t]
\begin{minipage}{1\linewidth}
\centering
\includegraphics[width=1\textwidth]{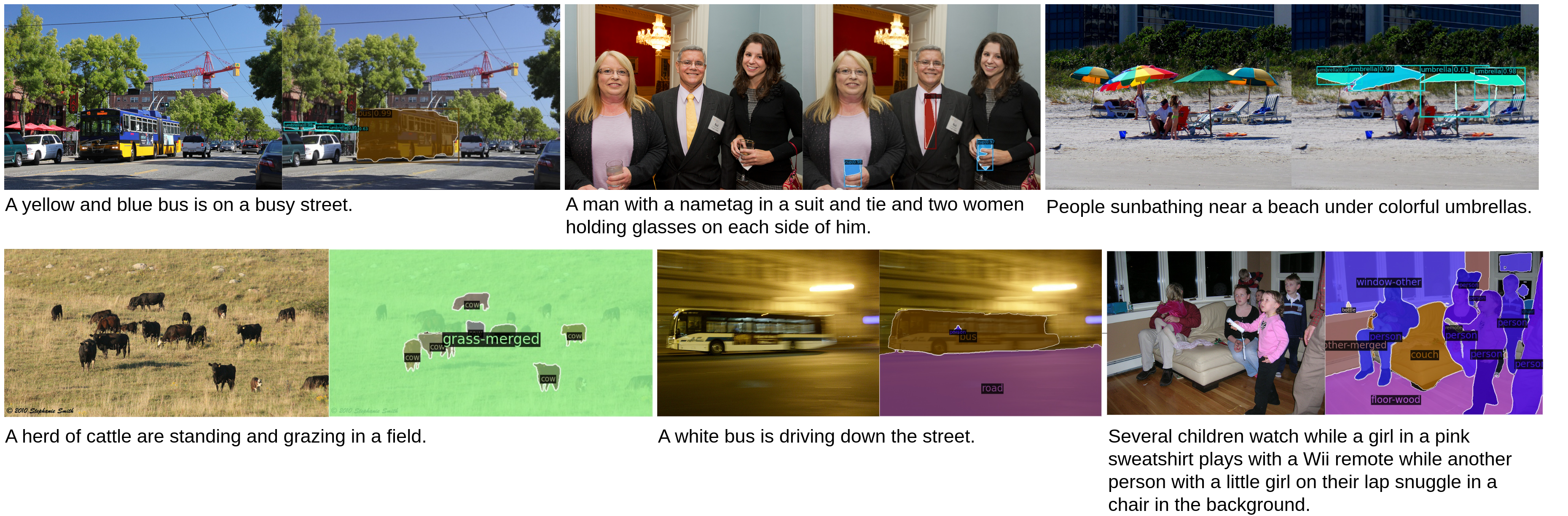}
\end{minipage}
\caption{Additional qualitative results for OVIS and OSPS. The first and second rows correspond to OVIS and OSPS examples, respectively. The captions below the images are pseudo captions generated by LLaVA.}
\label{fig:qualitative}
\end{figure*}

\vspace{1mm}
\noindent
\textbf{Qualitative Results}. 
Figure~\ref{fig:result_ovis} presents qualitative comparisons between the proposed method and the previous state-of-the-art method, CGG~\citep{wu2023betrayed}, on open-vocabulary instance segmentation. In the first row, our method successfully segments knives belonging to unknown classes, whereas CGG~\citep{wu2023betrayed} fails to detect them. Additionally, CGG~\citep{wu2023betrayed} incorrectly identifies a cup as an umbrella. In the second row, CGG~\citep{wu2023betrayed} misclassifies computer mice and laptops as keyboards, while our method accurately predicts a keyboard from novel categories. Furthermore, our method correctly segments the couch in the third row and classifies the cows in the last row.

Figure~\ref{fig:result_osps} presents qualitative comparisons between the proposed method and the previous state-of-the-art method, CGG~\citep{wu2023betrayed}, on open-set panoptic segmentation. In the first column, our method accurately segments the road, whereas CGG misclassifies it as grass. Moreover, our method more precisely identifies bananas in the second column and better distinguishes the stuff category in the last column.

Figure~\ref{fig:qualitative} presents additional qualitative results of the proposed MCCF framework. The first row shows open-vocabulary instance segmentation (OVIS) examples, while the second row shows open-set panoptic segmentation (OSPS) examples. The captions shown below each example are pseudo captions generated by LLaVA. These examples illustrate that MCCF can localize and segment objects described in the generated captions, demonstrating the usefulness of LLaVA-generated pseudo captions as multimodal supervision for both instance-level and panoptic segmentation settings.

\begin{figure*}[!t]
\centering
\begin{minipage}{0.25\linewidth}
\centering 
\includegraphics[width=1\textwidth,height=0.15\textheight]{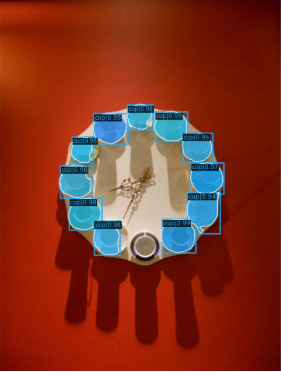}
\end{minipage}
\begin{minipage}{0.25\linewidth}
\centering 
\includegraphics[width=1\textwidth,height=0.15\textheight]{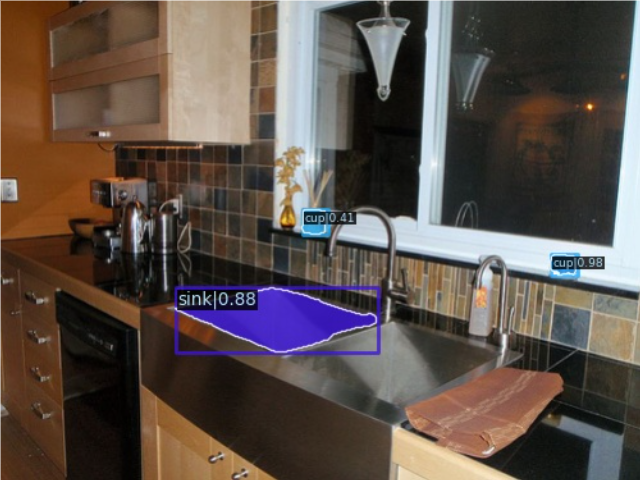}
\end{minipage}
\begin{minipage}{0.25\linewidth}
\centering 
\includegraphics[width=1\textwidth,height=0.15\textheight]{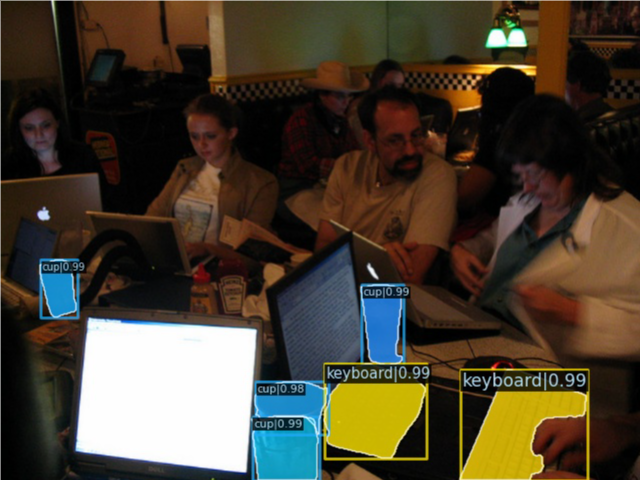}
\end{minipage}
\caption{Failure cases in OVIS. The examples show typical errors caused by over-segmentation, cluttered scenes, partial occlusion, low illumination, and ambiguous object boundaries.}
\label{fig:failure_case}
\end{figure*}

We present representative failure cases in Figure~\ref{fig:failure_case} to further analyze the limitations of the proposed pseudo-labeling strategy. In cluttered scenes, the model may also produce inaccurate masks or confuse neighboring objects, such as sinks, cups, and keyboards, due to partial occlusion, low illumination, or ambiguous object boundaries. Addressing these failure cases with more robust pseudo-label validation and boundary refinement remains an important direction for future work.

\begin{table*}[!t]
\centering
\caption{Ablation study on open-vocabulary instance segmentation.}
\label{tab:tbl_ablation_ovis}
\centering
\small
\begin{tabular}{>{\centering}m{0.16\textwidth}| >{\centering}m{0.1\textwidth}| >{\centering}m{0.1\textwidth}| >{\centering}m{0.1\textwidth}| >{\centering}m{0.13\textwidth}|  >{\centering\arraybackslash}m{0.13\textwidth}}
\toprule
\multicolumn{4}{c|}{Method} & \multicolumn{2}{c}{Novel AP} \\
\cmidrule(lr){1-4} \cmidrule(lr){5-6}
Baseline (CGG$^\dagger$) & $\calL_{gr}$ & $\calL_{cons}$ & $\calL_{recon}$ &Constrained & Generalized \\
\midrule
\checkmark & & & & 45.1 & 43.6 \\
\checkmark & \checkmark & & & 49.5 & 47.8 \\
\checkmark & \checkmark & \checkmark & & 50.9 & 49.3 \\
\checkmark & \checkmark & \checkmark & \checkmark & \textbf{51.6} & \textbf{50.4} \\
\bottomrule
\end{tabular}
\end{table*}

\subsection{Ablation Studies}
In Table~\ref{tab:tbl_ablation_ovis}, we present an ablation study evaluating the effectiveness of each proposed objective under the open-vocabulary instance segmentation setting. We begin with a CGG-based baseline trained using the same Grounded-SAM-based novel-class pseudo-mask annotations as our method, which corresponds to CGG$^\dagger$ in Table~\ref{tab:ovis}. This baseline retains the original CGG losses and achieves AP scores of 45.1 and 43.6 for novel categories under the constrained and generalized settings, respectively. Replacing the original CGG grounding loss with our extended grounding loss $\calL_{gr}$ in Eq.~(\ref{loss:grounding}) improves the performance to 49.5 and 47.8. Adding the semantic consistency loss $\calL_{cons}$ further enhances the results to 50.9 and 49.3. Finally, introducing the generative caption reconstruction loss $\calL_{recon}$ yields the best performance, achieving AP scores of 51.6 and 50.4 under the constrained and generalized settings, respectively.

We further analyze the effect of CLIP-based synonym filtering on the full MCCF model in Table~\ref{tab:synonym_filtering}. When CLIP-based filtering is removed, LLaVA-generated synonyms are used without visual verification, and the model achieves 47.1 and 44.0 AP on novel categories under the constrained and generalized settings, respectively. In contrast, applying CLIP-based filtering improves the performance to 51.6 and 50.4 AP, demonstrating that visually grounded synonym filtering effectively suppresses noisy language candidates and improves open-vocabulary generalization.

\begin{table}[!t]
\centering
\caption{Effect of CLIP-based synonym filtering on open-vocabulary instance segmentation.}
\label{tab:synonym_filtering}
\small
\begin{tabular}{>{\centering}m{0.2\textwidth}| > {\centering}m{0.1\textwidth}|  >{\centering\arraybackslash}m{0.1\textwidth}}
\toprule
Method & Constrained Novel & Generalized Novel \\
\midrule
MCCF w/o CLIP filtering & 47.1 & 44.0 \\
MCCF w/ CLIP filtering & \textbf{51.6} & \textbf{50.4} \\
\bottomrule
\end{tabular}
\end{table}

\noindent
\textbf{Computational Overhead.}
We further analyze the inference-time computational overhead of the proposed method. Table~\ref{tab:overhead} compares the CGG baseline and our model in terms of learnable parameters and GFLOPs. Our model increases the number of parameters from 35.6M to 38.4M and GFLOPs from 227.5 to 232.9, corresponding to relative increases of 7.9\% and 2.4\%, respectively. These results show that the proposed method introduces only modest additional inference cost. It is also worth noting that the auxiliary pseudo-label generation components, including Grounded SAM, LLaVA, and CLIP-based synonym filtering, are used only during pseudo-label construction and training, and are not required during inference.

\begin{table}[!t]
\centering
\caption{Inference-time computational overhead comparison with the CGG baseline.}
\label{tab:overhead}
\small
\begin{tabular}{>{\centering}m{0.2\textwidth}| >{\centering}m{0.1\textwidth}|  >{\centering\arraybackslash}m{0.1\textwidth}}
\toprule
Method & Parameters (M) & GFLOPs \\
\midrule
CGG~\citep{wu2023betrayed} & 35.6 & 227.5 \\
MCCF (Ours) & 38.4 & 232.9 \\
\bottomrule
\end{tabular}
\end{table}
 
\section{Conclusion}\label{sec:conclusion}
This paper presents a multimodal framework for open-vocabulary instance segmentation (OVIS) and open-set panoptic segmentation (OSPS), designed to enhance visual-textual alignment and generalization to unseen categories without requiring manual annotations. In our target-vocabulary-assisted pseudo-labeling setting, target novel-category names are used as prompts during pseudo-mask generation. Under this protocol, the proposed approach leverages pre-trained vision-language models to generate pseudo segmentation masks, descriptive captions, and semantically aligned synonym sets, thereby providing rich multimodal supervision. In addition, we introduced a semantic consistency loss and an extended grounding loss that incorporate both predefined category names and visually grounded synonyms to improve robustness to vocabulary variations. A GPT-based generative caption reconstruction loss was further proposed to strengthen fine-grained visual-textual reasoning by reconstructing masked captions conditioned on visual features. Extensive experiments on the COCO dataset demonstrate that our method consistently outperforms previous state-of-the-art approaches on both OVIS and OSPS benchmarks under the target-vocabulary-assisted protocol, validating the effectiveness of our multimodal pseudo-labeling and alignment strategy.

Rather than replacing existing segmentation architectures with a new decoder or backbone, our results show that open-vocabulary instance and open-set panoptic segmentation can be substantially improved by enriching training supervision with target-vocabulary-assisted multimodal pseudo-labels and synonym-aware visual-textual alignment objectives.

\vspace{1mm}
\noindent
\textbf{Limitation and Future Work}.
While the proposed framework demonstrates strong performance, computational resource constraints prevented training on large-scale datasets such as LVIS~\citep{gupta2019lvis} and Open Images~\citep{kuznetsova2020open}. Future work will explore large-scale training and computationally efficient learning strategies to further enhance open-vocabulary generalization and scalability.

\section*{Declaration of competing interest}
The authors declare that they have no known competing financial interests or personal relationships that could have appeared to influence the work reported in this paper.

\section*{Acknowledgments}
This research was supported in part by the National Research Foundation of Korea (NRF) grant funded by the Korea government(MSIT) (No. RS-2023-00252434).

\footnotesize
\bibliographystyle{elsarticle-harv}
\bibliography{egbib}

\end{document}